\documentclass{article}
\pdfoutput=1

\usepackage{arxiv}

\usepackage[utf8]{inputenc} 
\usepackage[T1]{fontenc}    
\usepackage{hyperref}       
\usepackage{url}            
\usepackage{booktabs}       
\usepackage{amsfonts}       
\usepackage{nicefrac}       
\usepackage{microtype}      
\usepackage{natbib}
\usepackage{multirow}
\usepackage{graphicx}

\usepackage{amsmath} 
\usepackage{amssymb}
\usepackage{array}
\usepackage{hhline}

\usepackage{xspace}

\usepackage[disable]{todonotes}
\newcommand{\rempz}[1]{\todo[color=orange]{{\scriptsize pz: #1\par}}\xspace}
\newcommand{\rempzi}[1]{\todo[color=orange,inline]{{\scriptsize pz: #1\par}}}

\newcommand*\samethanks[1][\value{footnote}]{\footnotemark[#1]}
\title{Cross-lingual Contextual Word Embeddings Mapping with Multi-sense Words in Mind}

\author{
Zheng Zhang\thanks{Equal contribution} \\
LIMSI, CNRS, \\ 
LRI, Univ.\ Paris-Sud, CNRS,\\
Université Paris-Saclay \\
Orsay, France \\
\texttt{zheng.zhang@limsi.fr} \\
\And
Ruiqing Yin\samethanks \\
LIMSI, CNRS, \\
Université Paris-Saclay \\
Orsay, France \\
\texttt{ruiqing.yin@limsi.fr} \\
\AND 
Jun Zhu\samethanks \\
CentraleSupélec \\
Université Paris-Saclay \\
Gif-sur-Yvette, France \\
\texttt{jun.zhu@centralesupelec.fr} \\
\And
Pierre Zweigenbaum \\
LIMSI, CNRS, \\
Université Paris-Saclay \\
Orsay, France \\
\texttt{pz@limsi.fr} \\
}

\begin{document}

\maketitle

\begin{abstract}
Recent work in cross-lingual contextual word embedding learning cannot handle multi-sense words well. In this work, we explore the characteristics of contextual word embeddings and show the link between contextual word embeddings and word senses. We propose two improving solutions by considering contextual multi-sense word embeddings as noise (removal) and by generating cluster level average anchor embeddings for contextual multi-sense word embeddings (replacement). Experiments show that our solutions can improve the supervised contextual word embeddings alignment for multi-sense words in a microscopic perspective without hurting the macroscopic performance on the bilingual lexicon induction task. For unsupervised alignment, our methods significantly improve the performance on the bilingual lexicon induction task for more than 10 points.
\end{abstract}


\section{Introduction}
Cross-lingual word embeddings (CLWEs), vector representations of words in multiple languages, are crucial to Natural Language Processing (NLP) tasks that are applied in multilingual scenarios, such as document classification, dependency parsing, POS tagging, named entity recognition, super-sense tagging, semantic parsing, discourse parsing, dialog state tracking, entity linking, 
sentiment analysis and machine translation~\citep{ruder2017survey}.

Cross-lingual word embedding learning models can be categorized into three groups based on when alignment data is used: corpus preparation, training and post-training.
For post-training models, research about the mapping
of state-of-the-art pre-trained monolingual word embeddings across different languages~\citep{mikolov2013efficient,joulin-etal-2017-bag,peters-etal-2018-deep,devlin-etal-2019-bert} keeps evolving with the progress of monolingual word embedding learning~\citep{mikolov2013exploiting,conneau2017word,lefever2009semeval,schuster-etal-2019-cross}. 

With the most recent progress of word embeddings learning by using pre-trained 
language representation models such as ELMo~\citep{peters-etal-2018-deep}, BERT~\citep{devlin-etal-2019-bert} and XLNet~\citep{yang2019xlnet}. Word embeddings move from context-independent to contextual representations.
\citet{peters-etal-2018-deep} have shown that contextual word embeddings have a richer semantic and syntactic representation.
For consistency and simplicity, we define two kinds of representations as word type embedding and token embedding.\\
\textbf{Word type embedding} Context-independent embedding of each word. Only one embedding is created for each distinct word in the training corpus.\\
\textbf{Token embedding} Contextual word embedding of each token. A token is one of the occurrences of a word (type) in a text, its embedding depends on its context. As a result, a word in the training corpus receives as many embeddings as its occurrences in that corpus.

Despite many advantages of token embeddings, mapping independently pre-trained token embeddings across languages is challenging: most existing word embeddings and cross-lingual mapping algorithms are based upon word type embeddings.
How to apply previous cross-lingual word embedding mapping algorithms to multi-sense word embeddings remains unclear.


\citet{schuster-etal-2019-cross} proposed the current state-of-the-art solution to this problem by conflating the multiple token embeddings of one word type into one context-independent embedding \emph{anchor}, which enables word-type-based cross-lingual word embedding learning algorithms to apply to token embeddings.
In their paper, the conflation of token embeddings is simply obtained by averaging them.

Although experiments show that this simple average anchor calculation is effective for cross-lingual token embeddings mapping, i.e. it obtained a better score on dependency parsing tasks than the previous state-of-the-art method, we believe there is still room for improvement, especially for multi-sense words.

\citet{schuster-etal-2019-cross} found that token embeddings for each word are well separated like clouds, and the token embeddings of a multi-sense word may also be separated according to different word senses inside each token embedding cloud.

Based on these findings, we argue that averaging is not an optimal choice for multi-sense word anchor calculation, which directly influences cross-lingual token embeddings learning.

\begin{itemize}
\item 
For the supervised mapping methods~\citep{mikolov2013exploiting,xing2015normalized}, the average anchor of a multi-sense word depends on the frequency of the token embeddings of each word sense. Besides, as each translation pair containing multi-sense words in the supervision dictionary may only cover one sense at one time, using only one anchor for each multi-sense word may not correspond to mono-sense based translation pairs.
\item
For the unsupervised cross-lingual word embedding learning model MUSE~\citep{conneau2017word}, because a multi-sense word may not have a translation word that would exactly have all its senses, the average anchor of that word may not find a corresponding average anchor embedding in the target language.
\end{itemize}

\paragraph{Our contributions}
The main contributions of this paper are 
the following:
\begin{itemize}
	\item
	Analyze the geometric distribution of token embeddings of multi-sense words, suggesting its relation to sense embeddings. 
	\item
	Using average anchor embeddings for both supervised and unsupervised cross-lingual word embedding learning models to show the existing problem.
	\item 
	Propose our solutions of treating multi-sense word anchor embeddings as noise and replacing word anchor embeddings with cluster-level average anchor embeddings.
\end{itemize}

\section{Related Work}
The learning method of~\citep{aldarmaki-diab-2019-context} relies on using parallel sentences either to generate a dynamic dictionary of token embeddings as the word-level alignment data or to calculate sentence embeddings as the sentence-level alignment data. \citet{schuster-etal-2019-cross} proposed to conflate the token embeddings for each word into one anchor embedding so as to apply previous cross-lingual word embedding learning algorithms, In the following, we focus on the solution of \citet{schuster-etal-2019-cross} as it does not need additional alignment data and it aims to connect all previous cross-lingual word embedding learning algorithms to the token embeddings field.

Below we introduce two cross-lingual word embedding learning methods along with their adaptations for token embeddings proposed by \citet{schuster-etal-2019-cross}.

\subsection{Supervised Mapping}
\label{sec:xing}
Supervised mapping methods aim to learn a linear mapping using the supervision of alignment data. \citet{mikolov2013exploiting} introduced a model that learns a linear transformation between word embeddings of different languages by minimizing the sum of squared Euclidean distances for the dictionary entries.
Based on this work, \citet{xing2015normalized} proposed an orthogonal transform to map the normalized word vectors in one or both languages under the constraint of the transformation being orthogonal because of two inconsistences in \citep{mikolov2013exploiting}:
\begin{itemize}
    \item During the skip-gram model training stage, the distance measurement is the inner product of word vectors according to the objective function while the cosine similarity is usually used for word embedding similarity calculation (e.g. for the WordSim-353 task).
    \item The objective function of the linear transformation learning step~\citep{mikolov2013exploiting} uses the Euler distance. But after mapping, the closeness of bilingual words is measured by the cosine similarity.
\end{itemize}

\citet{xing2015normalized}'s experiments showed that normalized word vectors have a better performance in the monolingual word similarity task WordSim-353 and that the proposed method performs significantly better in the word translation induction task than \citep{mikolov2013exploiting}. 

\paragraph{Adaptation for token embeddings}
Given a dictionary used for supervised cross-lingual context-independent word (word type) embedding learning, \citet{schuster-etal-2019-cross} proposed to generate average token embeddings anchors and to assign word anchor vectors to dictionay words.

\begin{equation}
    \label{eq:anchor1}
    \overline{e}_{i}=\mathbb{E}_{c}\left[\boldsymbol{e}_{i, c}\right]
\end{equation}

As shown in Equation~\ref{eq:anchor1}, the anchor embedding of word $i$ is defined as the average of token embeddings over a subset of the available unlabeled data, where ${e}_{i, c}$ is the token embedding of word $i$ in the context $c$.

\subsection{Unsupervised Mapping: MUSE}
\label{sec:muse}
MUSE (Multilingual Unsupervised and Supervised Embeddings) is a Generative Adversarial Net (GAN)-based method and open-source tool introduced by \citet{conneau2017word}. In their paper, a discriminator is trained to determine whether two word embeddings uniformly sampled from the $50,000$ most frequent words either come from the $WS$ (aligned source word embeddings, where $S$ is the source word embeddings and $W$ is the linear transformation matrix) or $T$ (target word embeddings) distributions. In the meantime, $W$ is trained to prevent the discriminator from doing so by making elements from these two different sources as similar as possible. Besides, they defined a similarity measure, Cross-domain Similarity Local Scaling (CSLS), that addresses the hubness problem (i.e., some points tend to be nearest neighbors of many points in high-dimensional spaces), and serves as the validation criterion for early stopping and hyper-parameter tuning.

\paragraph{Adaptation for token embeddings}
\citet{schuster-etal-2019-cross} also proposed another adaptation on top of the MUSE model~\citep{conneau2017word} by using anchor embeddings: as they did in the supervised case, anchor embeddings are assigned as the vector representations for words. Then they use them in the unsupervised MUSE model.

\section{Average Anchor Embedding for Multi-sense words}
\label{sec:avearging-problems}
Using the average for anchor calculation is based on two findings from \citet{schuster-etal-2019-cross}'s exploration of token embeddings:
\begin{enumerate}
    \item The clouds of token embeddings of each word are well separated.
    \item
      \begin{enumerate}
      \item The clouds of multi-sense words may be separated according to distinct senses.
      \item Although the distances between token embeddings and the averaged token embedding cloud center are slightly larger than in single-sense words, the token embeddings of multi-sense words \textit{still remain relatively close to their [...] anchor}. Because of this, the authors believe ``these anchors can still serve as a good approximation for learning alignments''.
      \end{enumerate}
\end{enumerate}

In our opinion however, there is no reason for the distance between token embeddings of distinct senses to be small.
Take the English word \textit{bank} as an example, which has multiple distinct senses including the meaning of a financial institution and the meaning of the river side. There is no reason why token embeddings related to the financial institution meaning should be close to token embeddings of the river side meaning.

We decided to investigate these claims by analyzing monolingual and aligned cross-lingual token embeddings.
Our empirical investigation is consistent with the first conclusion (1) and the first point of the second conclusion (2a), but disagrees with the second point of the second conclusion (2b).
Additionally, we attempt to explain why this second point is not likely to hold in principle.

\subsection{Token Embeddings}
\label{sec:token-embedding}

To show the difference of token embedding geometrical distributions between multi-sense words and single-sense words, 
we need a multi-sense word that is directly related to single-sense words. 
The English word \textit{lie} could be a good choice: the verb \textit{lie} has two distinct senses, and each sense has a different past tense: \emph{lied} (did not tell the truth) or \emph{lay} (was in a horizontal position, was located). Besides, the English word \textit{lie} can also be a noun, whose antonym is \textit{truth}. 

So we visualize the embeddings of the English word \textit{lie} along with its two past tenses \textit{lied} and \textit{lay} and one of its antonyms, \textit{truth}. 


\begin{figure}[htbp!] 
\centering    
\includegraphics[width=0.5\textwidth]{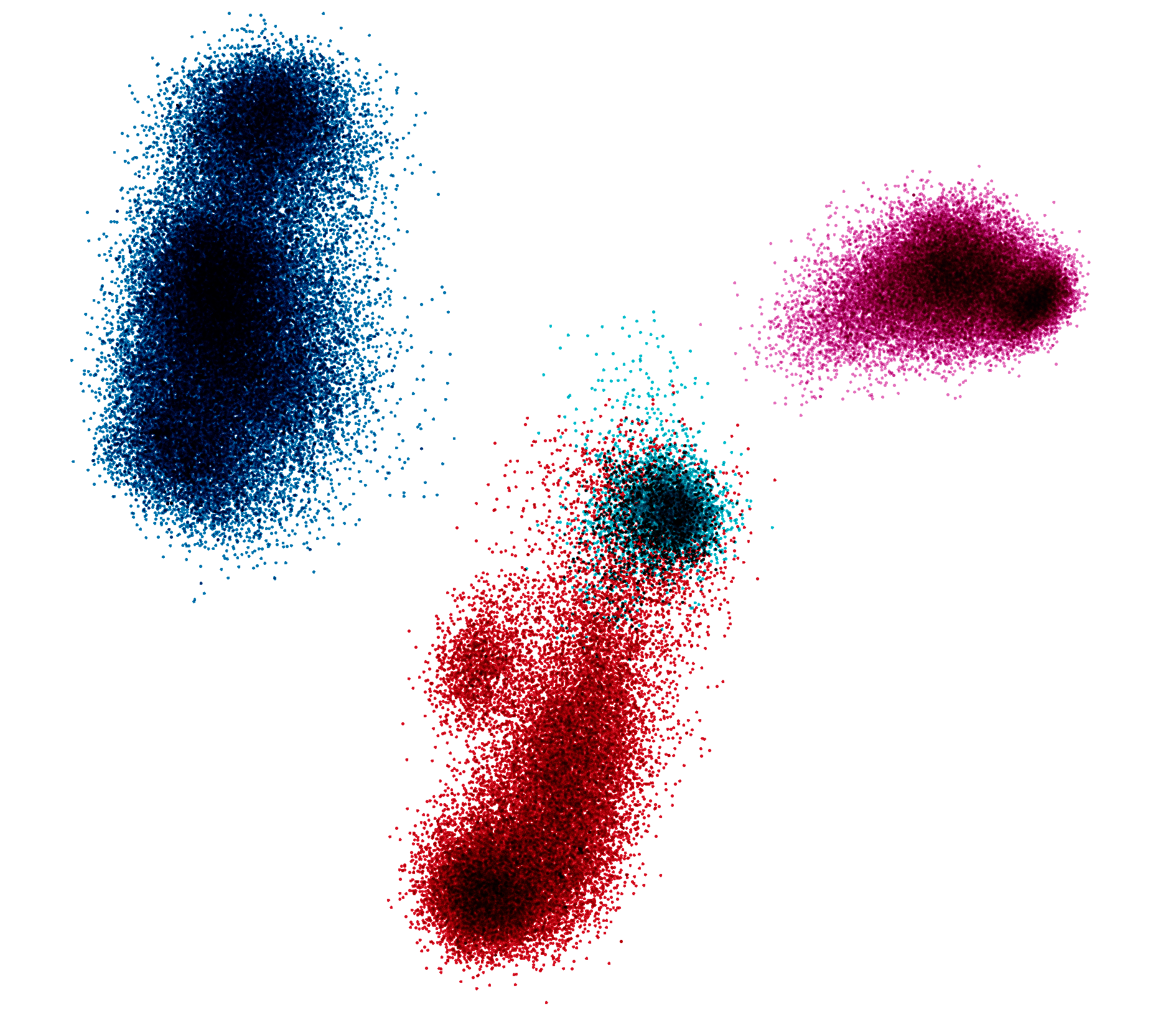}
\caption[Token embeddings of the English word \textit{lie} along with its two past tenses \textit{lied} and \textit{lay} and one of its antonyms \textit{truth}]{Token embeddings of the English word \textit{lie} (red points, bottom middle) along with its two past tenses \textit{lied} (light blue points, top middle) and \textit{lay} (dark blue points, top left) and one of its antonyms \textit{truth} (purple points, top right).}
\label{fig:lie}
\end{figure}

As shown in Figure~\ref{fig:lie}, we found that the point clouds of the single-sense words \textit{lied} and \textit{truth} are more concentrated than for the multi-sense word \textit{lie}.
The point cloud of the word \textit{lie} can be visually categorized into 3 clusters: one that overlaps the cloud of \textit{lied} in light blue, one at the bottom, and another one on the left.
By randomly selecting points and checking their corresponding sentences (Table~\ref{tab:lie-sent}) from each cluster, we found that the point clouds of the word \textit{lie} are separated according to its distinct senses.
Surprisingly, we also found that the point cloud of the word \textit{lay} is also visually separated into 2 parts. By checking the corresponding sentences, We found the bottom part is used as the past tense of the word \textit{lie}s and the top part is used as an adjective. 
 (Three corresponding sentences: \textit{In 1980, Mr. Greg Coffey was appointed the first \underline{lay} principal of the College.}, \textit{Conwell took up the post at an advanced age, and spent much of his time there feuding with the \underline{lay} trustees of his parishes, especially those of St. Mary's Church in Philadelphia.} and \textit{This includes a wide range of relationships, from monastic (monks and nuns), to mendicant (friars and sisters), apostolic, secular and \underline{lay} institutes.}). 


\begin{table}[htbp!]
\centering
\resizebox{1.0\textwidth}{!}{%
  \begin{tabular}{lp{10cm}p{3cm}}
    \hline
    \begin{tabular}[c]{@{}l@{}}Cluster\\ position\end{tabular} & Sentence & \begin{tabular}[c]{@{}l@{}}Semantic\\ category\end{tabular}\\
    \hline
    \multirow{7}{*}{overlapping} & \textit{Yutaka and Lady Elizabeth come to the hearing and \textbf{lie} to incriminate Oyuki.} & \multirow{7}{*}
    {\begin{tabular}[c]{@{}l@{}}[verb] to deliberately\\ say sth that is not true\end{tabular}}\\
    & \textit{As a result of his confession, prosecutors decided not to pursue a prosecution against the remaining 20 charges, and asked that they \textbf{lie} on file, in order to spare a jury the horror of having to watch graphic images and videos of child abuse since the 71 charges which Huckle admitted to would be sufficient for a lengthy sentence.} & \\
    \hline
    
    \multirow{5}{*}{bottom} & \textit{The city's prime locations \textbf{lie} within a radius of 6 km from Thammanam, making it thus a predominantly residential and small commercial area with basic facilities in and around the region.} & \multirow{5}{*}
    {\begin{tabular}[c]{@{}l@{}}[verb] to be in a\\ particular position\end{tabular}}\\
    & \textit{As of 2009, the most heavily trafficked segments of NY 31 \textbf{lie} in and around the city of Rochester.} & \\
    \hline
    
    \multirow{4}{*}{left} & \textit{James Murphy later admitted that this was entirely a \textbf{lie} on his part, and that he does not actually jog.} & \multirow{4}{*}
    {\begin{tabular}[c]{@{}l@{}}[noun] sth you say that\\ you know is not true\end{tabular}}\\
    & \textit{The dater then asks the suitors questions which they must answer while hooked up to a \textbf{lie} detector, nicknamed the "Trustbuster".} & \\
    \hline
  \end{tabular}
  }
  \caption{Corresponding sentences selected from each visual clusters of the token embeddings of the word \textit{lie}}
  \label{tab:lie-sent}
\end{table}

Similar findings can be found in the token embeddings of other words, in different languages and also in aligned cross-lingual embedding spaces. As suggested by~\citet{schuster-etal-2019-cross}'s conclusion, point clouds for each word are well separated (Conclusion~1). Besides, the point clouds of multi-sense words are also separated according to distinct senses (Conclusion~2a).

\subsection{Average Anchor Embeddings for Multi-sense Words}
\label{sec:average-anchor}
\begin{figure}[htbp!]
\centering
  \begin{tabular}{@{}cc@{}}
\includegraphics[width=.5\textwidth]{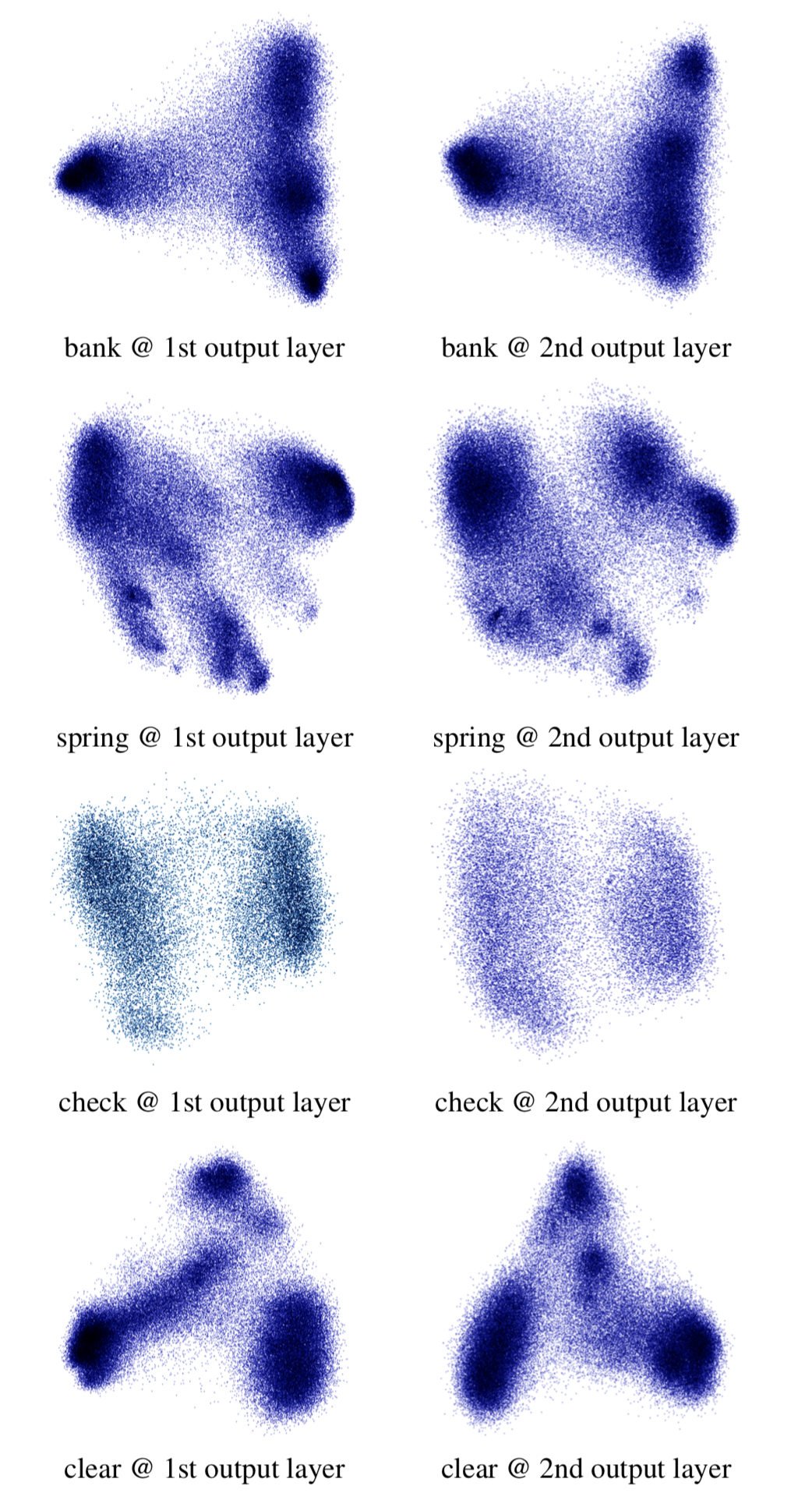} &
\includegraphics[width=.5\textwidth]{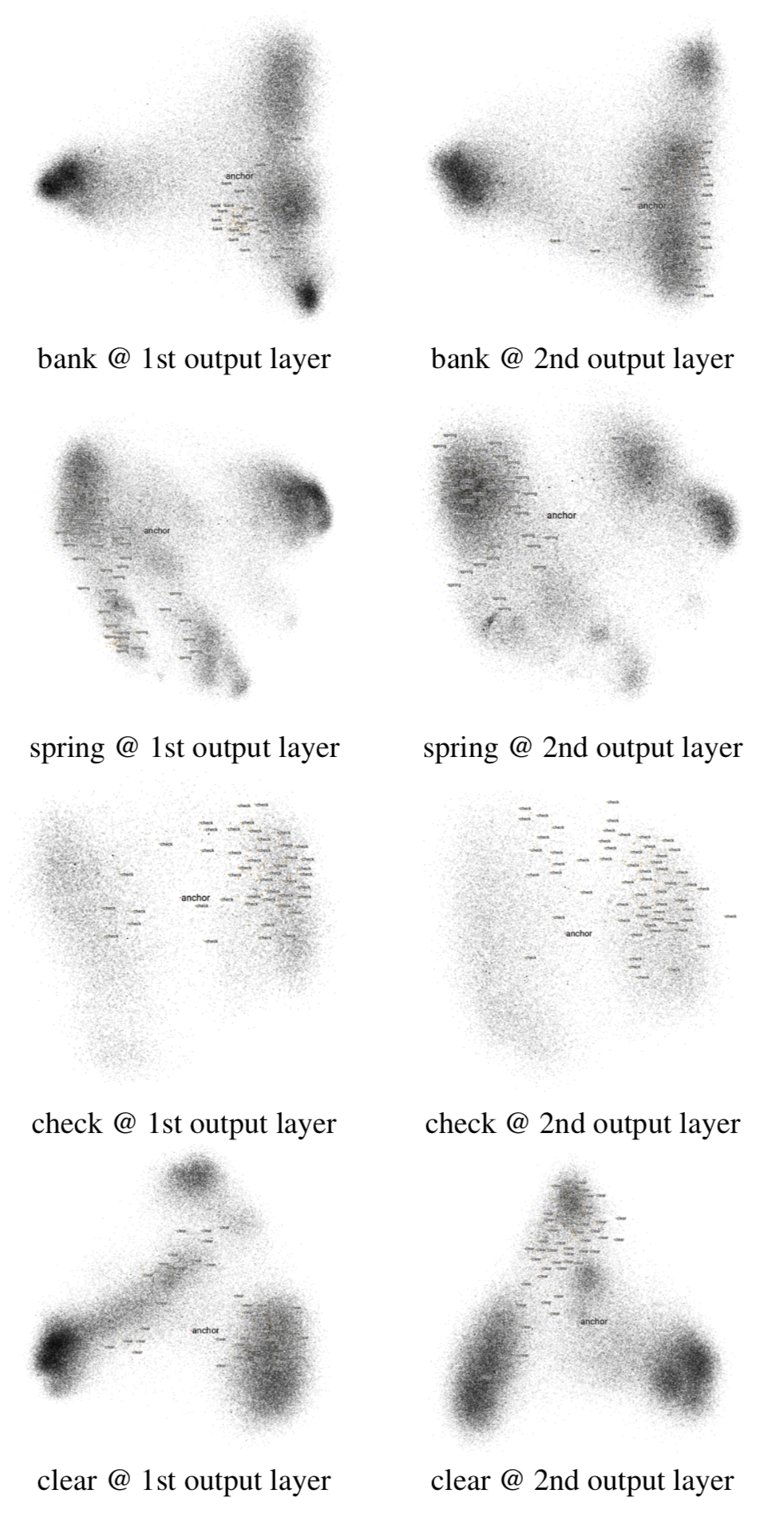} \\
\end{tabular}
  \caption[Token embeddings of English words \textit{bank}, \textit{spring}, \textit{check} and \textit{clear} generated from the first and second LSTM layers of the ELMo model,
  Anchor embeddings of English words \textit{bank}, \textit{spring}, \textit{check} and \textit{clear} and their 100 nearest token embeddings]
  {Token embeddings of English words \textit{bank}, \textit{spring}, \textit{check} and \textit{clear} generated from the first and second LSTM layers of the ELMo model.
  Labelling of the anchor embeddings (\emph{anchor}) of English words \textit{bank}, \textit{spring}, \textit{check} and \textit{clear} and of their 100 nearest token embeddings (\emph{bank}, \emph{spring}, etc.). Embeddings are generated from the first and the second LSTM layers of the ELMo model.}
  \label{fig:4-multisense-words}
\end{figure}

To analyze multi-sense word token embeddings and their average anchors in detail, we manually selected 4 multi-sense English words from the Wikipedia list of true homonyms
from different perspectives:
\begin{itemize}
    \item Distinct senses of the same part of speech (POS) (noun): \textbf{bank}-financial, \textbf{bank}-river, etc.; \textbf{spring}-season, \textbf{spring}-fountain, \textbf{spring}-coiled, etc.
    
    \item Distinct senses of different POS: \textbf{check}/Noun \textbf{check}/Verb;
 \textbf{clear}/Adj,  \textbf{clear}/Verb
    

\end{itemize}
\paragraph{Distribution of token embeddings for multi-sense words.}
We firstly calculate all the token embeddings of the selected words over the whole English Wikipedia.
We use the output of either the first or second LSTM layer of ELMo as input to the visualization (see Figure~\ref{fig:4-multisense-words}).

\paragraph{Position of anchor embeddings for multi-sense words.}
Besides the embeddings projection, we also calculate anchor embeddings for the selected multi-sense words. Then we label the 100 nearest neighbors of each anchor in the token embedding space (see the right side of Figure~\ref{fig:4-multisense-words}).
Note that all token embeddings are also present in that visualization, but only the top 100 are labeled with the word.

\paragraph{Context of token embeddings.}
Also, to verify that token embeddings are geometrically separated according to distinct senses, for each cluster in the point cloud of a multi-sense word, we randomly select two points (token embeddings) in this cluster and show their corresponding sentences (see Appendix). Note that we do not apply any clustering algorithm here, clusters are just recognized based on human judgment.

\paragraph{Observation.}
As shown in the right side of Figure~\ref{fig:4-multisense-words}, most of the 100 token embeddings nearest to the anchor embedding are located in only one of the word sense clusters. 
The anchor is pulled closer to the sense clusters that have more token embeddings because of the averaging, which causes the first problem for cross-lingual token embeddings mapping:\\
\textbf{Problem 1}  The anchor of a multi-sense word is biased by the frequency of the token embeddings of its senses.

\subsection{Muti-sense Words in Dictionaries for Supervised Mapping}
\label{sec:anchor-supervised}
The supervised model is trained on a bilingual dictionary of source-target words. Dictionaries are not always generated with attention paid to multi-sense words. When a dictionary contains incomplete translation pairs related to a multi-sense word, it may contribute inaccurate mapping supervision data.

Let us take as an example the English-French dictionary, containing 5,000 source words, used for the supervised baseline model in MUSE. We list in Table~\ref{tab:bank-in-dict} all translation pairs in that dictionary related to a common multi-sense word: \textit{bank}.

\begin{table}[htbp!]
  \centering\setlength{\tabcolsep}{4pt}
  \begin{tabular}{ll}
    \hline
    bank & banques\\
    bank & banque\\
    banks & banques\\
    banking & banques\\
    banking & banque\\
    banking & bancaire\\
    \hline
  \end{tabular}
  \caption[All translation pairs related to the multi-sense word \textit{bank} in the English-French dictionary used in MUSE for supervised mapping]{All translation pairs related to the multi-sense word \textit{bank} in the English-French dictionary used in MUSE for supervised mapping.}
\label{tab:bank-in-dict}
\end{table}

It is obvious that all translation pairs listed above are related to the \textit{financial institution} meaning of the word \textit{bank}. The other senses of bank, such as \textit{land at river's side}, are ignored. Similar cases can be found for other multi-sense words in the dictionary.\\

\textbf{Problem 2}  Because the average anchor for a multi-sense word can be considered as a general representation of all its distinct senses, using this for semantically incomplete translation pairs in a dictionary may lead to inaccurate mappings.

\subsection{Muti-sense Words for the Unsupervised Mapping in MUSE}
\label{sec:prob-muse}
The unsupervised mapping model in MUSE uses a GAN to learn a linear mapping between source and target embeddings without parallel supervision data. 
Based on the intuition that source and target embedding spaces should share a similar global geometric structure, in the best case, source words should be mapped to their corresponding translation words in target languages.

\textbf{Problem 3} For multi-sense words, translations that have exactly the same set of senses may not exist, e.g. for the English word \textit{bank}, there is no corresponding French word which has both the \textit{financial institution} (``banque'') and \textit{land at river's side} (``berge'', ``bord'', ``rive'', etc) senses.
Therefore a multi-sense word anchor may not have a corresponding point in the target language.

\section{Cross-lingual Token Embeddings Mapping with Multi-sense Words in Mind}
\label{sec:anchor-noise}
We propose below solutions to these problems for both supervised mapping and unsupervised mapping methods.

\subsection{Noise in Dictionary for Supervised Mapping}
We consider incomplete translation pairs of multi-sense words as noise in the supervision data (dictionary).
A simple but effective solution is to remove noise.
Here we propose two types of removal:
\begin{description}
    \item[Form-based removal:]
    remove translation pairs that contain the exact multi-sense words. For instance, given that the source word \textit{bank} is known to have multiple senses, \textit{bank banques} and \textit{bank banque} should be removed in Table~\ref{tab:bank-in-dict}.
    \item[Lemma-based removal:] remove translation pairs containing words having the same lemma as multi-sense words. In the \textit{bank} example, all 6 translation pairs in Table~\ref{tab:bank-in-dict} should be removed as \textit{bank}, \textit{banks}, and \textit{banking} have the same lemma. 
\end{description}

Note that we do not supply a part of speech (POS) tag to the lemmatizer as there is no context to analyze the POS for words in the translation pairs of the dictionary.


\subsection{Noisy Points for Unsupervised Mapping in MUSE}
As discussed before, the exact corresponding senses-to-senses translation of a multi-sense word may not exist in target languages, i.e. the average anchor for multi-sense words may not be correctly aligned to target embedding spaces.

In that context, we consider multi-sense word anchors as noise for the unsupervised mapping model in MUSE. So we remove all multi-sense word anchors
\rempz{There is some redundancy here wrt the previous subsection} from the independently pre-trained monolingual word embeddings used for training (We name this method \textbf{anchors removal} in Table~\ref{tab:eval-result}).

\subsection{Cluster-level Average Anchor Embeddings for Unsupervised Mapping in MUSE}
We apply the spectral clustering algorithm~\cite{wang2018speaker} to token embeddings of multi-sense words and calculate an average anchor embedding for each cluster.
Then for each multi-sense word, we replace its average anchor embedding with cluster-level average anchor embeddings. (We name this method \textbf{anchors replacement} in Table~\ref{tab:eval-result}.)

\section{Experiments}
\subsection{Token Embeddings}
\textbf{Pre-trained model} We use the same ELMo models
as in \citep{schuster-etal-2019-cross}, which are trained on Wikipedia dumps
with the default parameters of ELMo~\citep{peters-etal-2018-deep}.\\
\textbf{Corpus} The Wikipedia dumps we used for specific words analysis are the same as the training data for ELMo models.\\
\textbf{Lexicon induction evaluation} Following \citep{schuster-etal-2019-cross}, we use average anchors
to produce word translations to evaluate alignments.
For the clustering based method, we use cluster-level average anchors of multi-sense words.
Gold standard dictionaries are taken from the MUSE framework
and contain 1,500 distinct source words.

\subsection{Supervised Mapping}
\textbf{Dictionary} The baseline supervised linear mapping is calculated based on a dictionary of 5,000 distinct source words downloaded from the MUSE library.
\\
\textbf{Corpus for word occurrence embedding and anchor calculation}
We compute the average of token embeddings on a fraction (around 500MB, or 80 million words)
of English (/French) Wikipedia dumps as anchor vectors for the English (/French) words in dictionaries.

\subsubsection{Detailed Analysis about \textit{bank}}
\label{sec:exp-bank}
To obtain an intuitive understanding of how multi-sense words behave in supervised mapping methods, we start our supervised mapping experiment focusing on a common English multi-sense word \textit{bank}.\\
\textbf{2 dictionaries used for supervised linear mapping}
To analyze the influence of incomplete translation pairs about \textit{bank} in the dictionary, we generate two filtered dictionaries by removing translation pairs containing \textit{bank} (form-based removal: \textit{bank} $\Leftrightarrow$ \textit{banques} and \textit{bank} $\Leftrightarrow$ \textit{banque}) and by removing translation pairs having the same lemma as \textit{bank} (lemma-based removal: \textit{bank} $\Leftrightarrow$ \textit{banques}, \textit{bank} $\Leftrightarrow$ \textit{banque}, \textit{banks} $\Leftrightarrow$ \textit{banques}, \textit{banking} $\Leftrightarrow$ \textit{banques}, \textit{banking} $\Leftrightarrow$ \textit{banque}, and \textit{banking} $\Leftrightarrow$ \textit{bancaire}).

For token embeddings visualization, we compute token embeddings of the English word \textit{bank} and of its French translations (i.e. ``banque'', ``bord'', ``rive'', and ``berge'', according to the Collins English-French Dictionary
and WordReference.com%
) over around $500MB$ English and French corpora.

\subsubsection{Removal of English and (or) French Multi-sense Words}
Based on the Wikipedia list of English homonyms%
, we generate two dictionaries by form-based removal and lemma-based removal.
The original dictionary has $9496$ valid translation pairs, the form-based removal dictionary has $9161$ valid translation pairs and the lemma-based removal dictionary has $9076$.

For French, we generate four dictionaries by form-based removal and lemma-based removal based on two French polyseme lists.\rempz{Do they overlap? Say a word about their properties.}
The form-based removal dictionaries have $9416$ and $9331$ valid translation pairs and the lemma-based removal dictionaries have $9370$ and $9226$ based on two lists respectively.

Furthermore, we also tried to remove both English and French Multi-sense Words by form-based removal and lemma-based removal.

\subsection{Unsupervised Mapping}
We calculate token embeddings for the 50,000\rempz{Say a word to justify this number.} most frequent words in English and in the target language. For frequent words selection, we follow the word order in FastText pre-trained word vectors%
, which are sorted in descending order of frequency.
The corpus used for anchor calculation and also the multi-sense word lists are the same as those used for supervised mapping.

To apply the spectral clustering algorithm to multi-sense word token embeddings, we calculate the frequency of token embeddings first. 
If it is less than 160, we keep the original average anchor embedding. If it is larger than 10,000\rempz{Say a word to justify this number.}, we randomly sample a subset of 10,000 token embeddings and then apply the clustering algorithm to it.

\subsection{Set-up for Embedding Visualization}
Embedding Projector\footnote{http://projector.tensorflow.org} has been used for data visualization.
We generate two 2-D graphs for each selected polysemy (or polysemies) by selecting PCA (Principal Component Analysis) for dimensionality reduction and Sphereize data \textit{(The data is normalized by shifting each point by the [coordinates of the] centroid and making it unit [length])}
for data normalization.

Note that PCA is approximate in the Embedding Projector, i.e., \textit{for fast results, the data was sampled to 50,000 points and randomly projected down to 200 dimensions.}
As token embeddings generated by ELMo have 1024 dimensions, the embeddings used for visualization were randomly projected down to 200 dimensions.

\section{Results}

\begin{table*}[htp]
  \centering
  \setlength{\tabcolsep}{2pt}
  \begin{tabular*}{\textwidth}{l@{\extracolsep{\fill}}*{12}{c}}
    \hline
    \multirow{3}{*}{Alignment} & \multicolumn{6}{c}{1st LSTM output layer} & \multicolumn{6}{c}{2nd LSTM output layer} \\
    \cline{2-7} \cline{8-13}
    & \multicolumn{3}{c}{nn} & \multicolumn{3}{c}{csls\_knn\_10} & \multicolumn{3}{c}{nn} & \multicolumn{3}{c}{csls\_knn\_10}\\
    \cline{2-4} \cline{5-7} \cline{8-10} \cline{11-13}
    & P@1 & P@5 & P@10 & P@1 & P@5 & P@10 & P@1 & P@5 & P@10 & P@1 & P@5 & P@10\\
    \hline
    \multicolumn{13}{c}{(a) Supervised Mapping}\\
    \hline
    Baseline & \textbf{55.20} & 73.85 & 80.11 & 68.48 & 84.65 & 88.78 &
    \textbf{55.95} & \textbf{73.57} & 79.49 & \textbf{67.17} & 82.31 & 86.79\\
    Form-based removal (en) & 54.99 & 74.19 & 79.63 & \textbf{68.55} & \textbf{85.13} & 88.58 &
    55.33 & 73.43 & 79.22 & 66.96 & \textbf{82.59} & 86.51\\
    Form-based removal (fr-1) & 54.85 & \textbf{74.26} & 79.77 & \textbf{68.55} & 84.86 & 88.92 &
    55.88 & 73.50 & \textbf{79.63} & 66.90 & 82.11 & 86.79\\
    Form-based removal (fr-2)& 54.85 & 73.85 & \textbf{80.32} & 68.27 & 84.65 & 88.71 &
    55.81 & \textbf{73.57} & \textbf{79.63} & 67.10 & 82.31 & 86.85\\
    Lemma-based removal (en) & 55.06 & 74.05 & 79.83 & 68.41 & 85.07 & 88.64 &
    55.33 & 73.30 & 79.15 & 66.62 & 82.38 & 86.58\\
    Lemma-based removal (fr-1) & 54.92 & 74.19 & 79.83 & 68.07 & 84.79 & \textbf{89.13} &
    55.82 & \textbf{73.57} & 79.56 & 66.83 & 82.17 & 86.79\\
    Lemma-based removal (fr-2) & 54.85 & 73.57 & 80.11 & 68.41 & 84.72 & 88.71 &
    55.74 & \textbf{73.57} & \textbf{79.63} & 66.83 & 82.38 & \textbf{86.99}\\
[1ex]
    
    
    \hline
    \multicolumn{13}{c}{(b) Unsupervised Mapping}\\
    \hline
    Baseline & 42.81 & 62.70 & 67.72 & 48.11 & 69.99 & 74.54 &
    35.58 & 49.90 & 56.64 & 42.60 & 62.42 & 68.62\\
    Anchors removal (en) & 52.44 & 67.38 & 72.06 & 57.88 & 73.43 & 77.22 &
    \multicolumn{6}{c}{No convergence}\\
    Anchors removal (fr-1) & 48.59 & 63.11 & 67.65 & 53.68 & 69.37 & 72.61 &
    \textbf{47.69} & \textbf{61.73} & \textbf{67.45} & \textbf{53.34} & \textbf{70.27} & \textbf{76.05}\\
    Anchors removal (fr-2) &  45.97 & 60.16 & 64.30 & 50.52 & 65.33 & 69.81 &
    \multicolumn{6}{c}{No convergence}\\
    Anchors removal (en \& fr-1) & \multicolumn{6}{c}{No convergence} &
    36.89 & 51.41 & 57.47 & 41.77 & 60.70 & 67.72\\
    Anchors removal (en \& fr-2) & 51.96 & 68.44 & 73.12 & 58.17 & 75.53 & 79.53 &
    33.43 & 45.83 & 50.59 & 39.83 & 53.55 & 59.27\\
    Anchors replacement (en) & \textbf{54.71} & \textbf{70.54} & \textbf{75.02} & \textbf{60.98} & \textbf{78.32} & \textbf{82.38} & \multicolumn{6}{c}{No convergence}
    \\
    \hline
  \end{tabular*}
  \rempzi{Note that information on the supervised and unsupervised methods could be moved to the table heading rows (Supervised Mapping, Unsupervised Mapping) to save space.}
  \caption{Precision at $k=1, 5, 10$ of bilingual lexicon induction from the aligned cross-lingual embeddings.}
  \label{tab:eval-result}
\end{table*}

\subsection{Visualization of the Token Embeddings of \textit{bank}}
\label{sec:results-bank}
Experiment results are shown in three figures presented below, in which dark blue points represent the English word \textit{bank}, light blue points are token embeddings for the French word \textit{banque}, and the French words \textit{berge}, \textit{bord}, \textit{rive} are in green, red and pink colors respectively.

\begin{figure}[htbp!] 
\centering    
\begin{tabular}{@{}ccc@{}}
\includegraphics[width=0.35\textwidth]{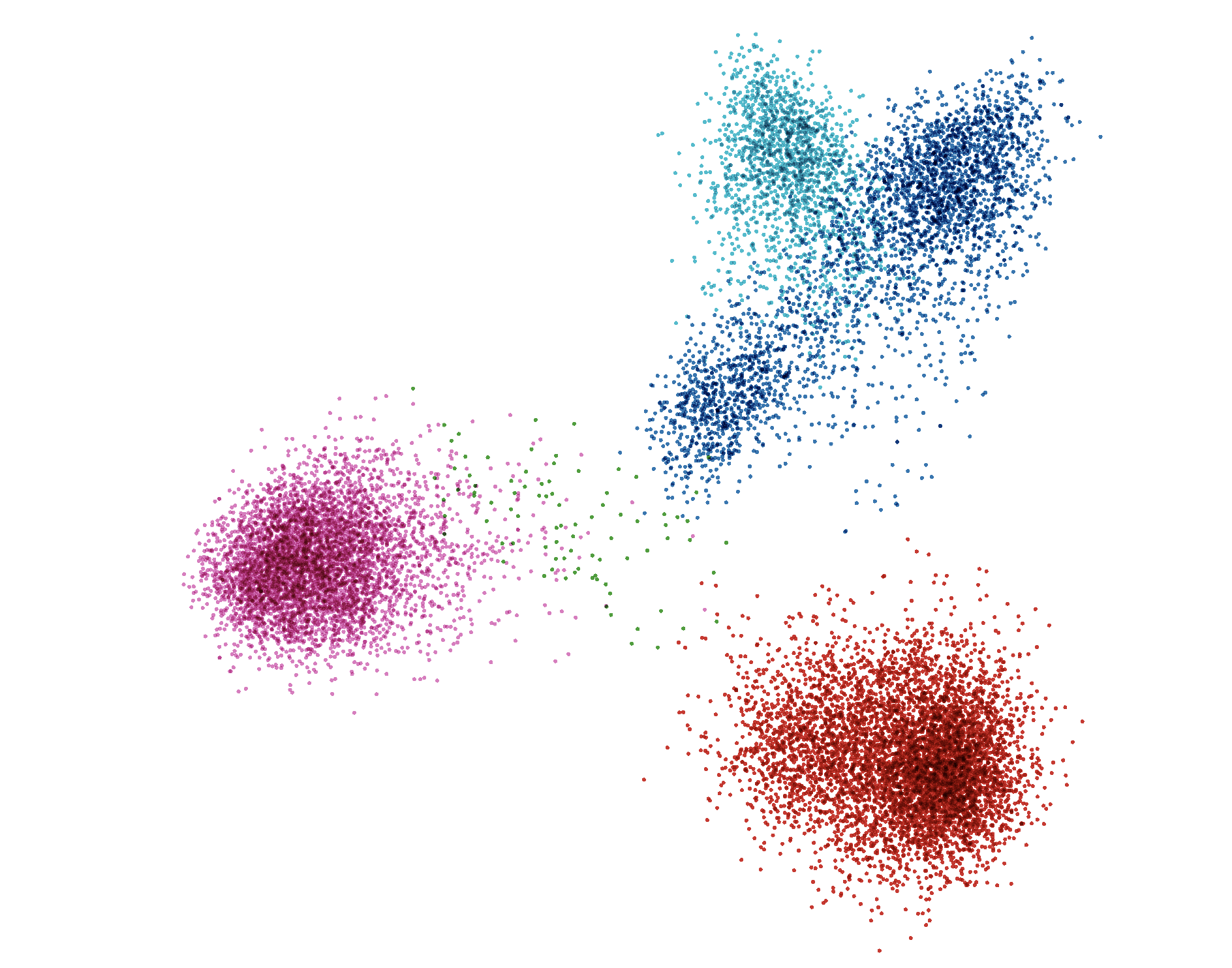} &
\includegraphics[width=0.29\textwidth]{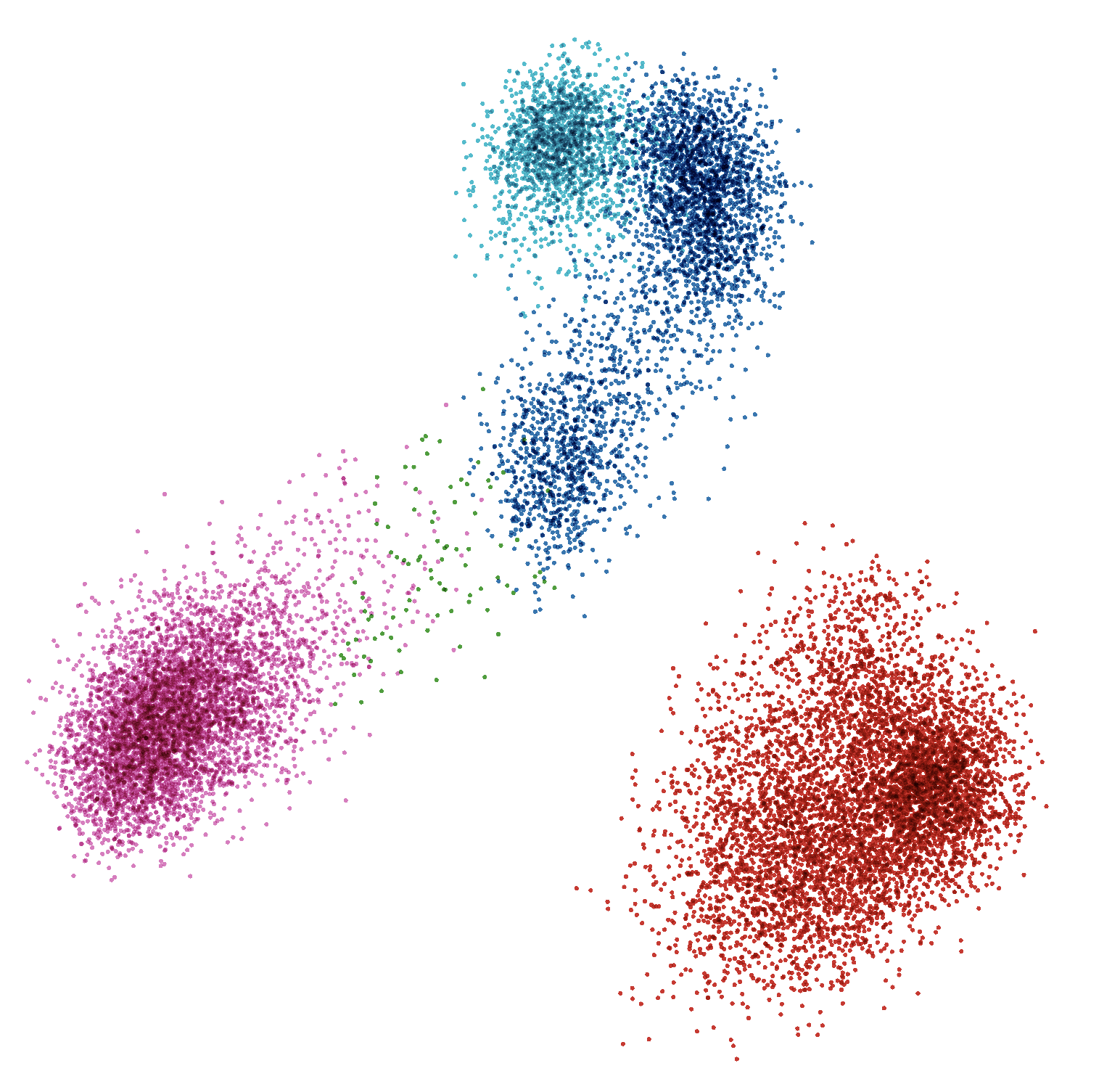} &
\includegraphics[width=0.34\textwidth]{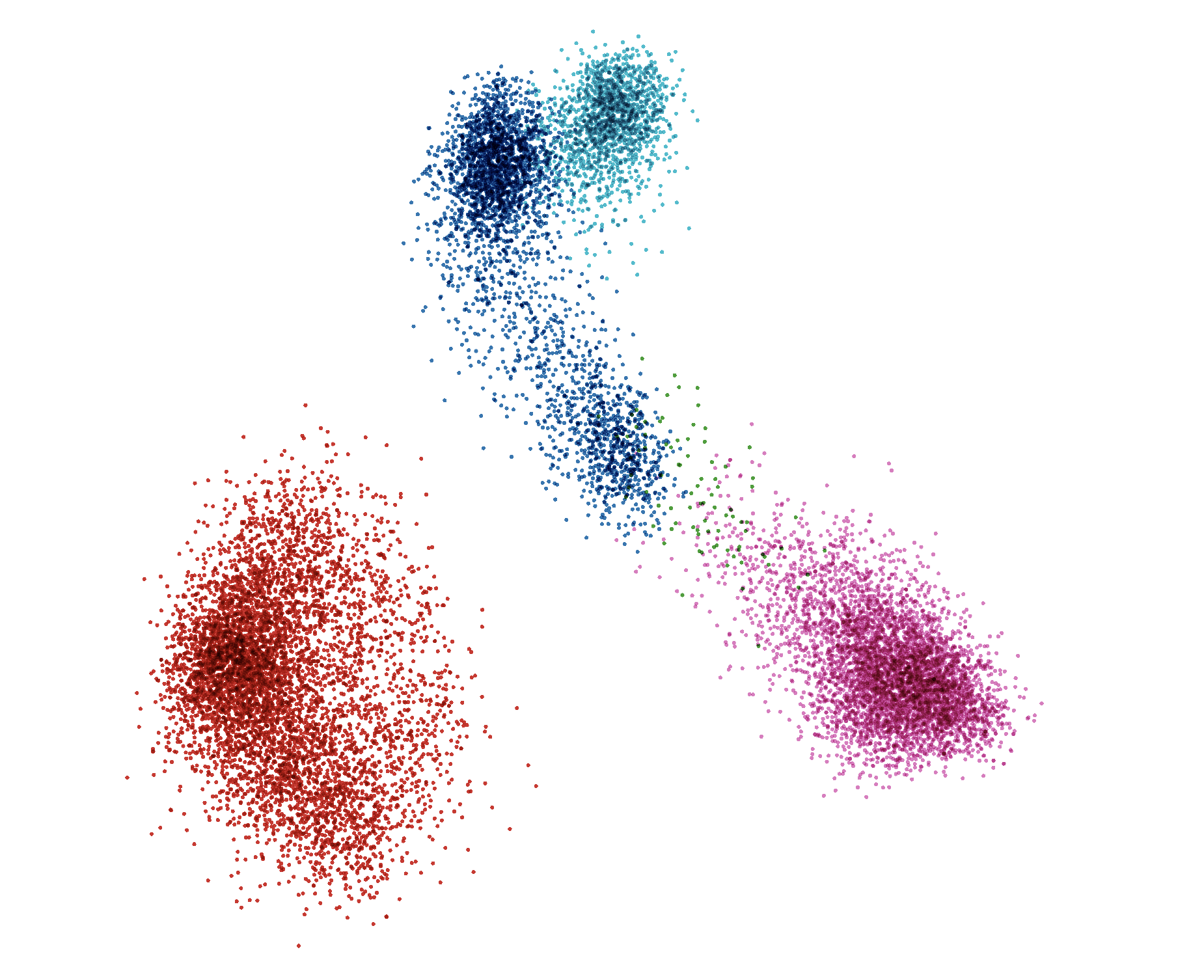}\\
\end{tabular}

\caption[Aligned token embeddings for the English word \textit{bank} and French words \textit{banque}, \textit{berge}, \textit{bord} and \textit{rive} after removing translation pairs having the same lemma of \textit{bank}]{Aligned token embeddings for the English word \textit{bank} (in dark blue) and French words \textit{banque} (in light blue), \textit{berge} (in green), \textit{bord} (in red) and \textit{rive} (in pink).
Baseline alignment shown on the left, alignment after removing translation pairs having the same form as \textit{bank} shown in the middle and alignment after removing translation pairs having the same lemma as \textit{bank} shown on the right.}
\label{fig:bank-banque_berge_bord_rive_00}
\end{figure}

As shown in Figure~\ref{fig:bank-banque_berge_bord_rive_00}, in the baseline aligned embedding space, the point cloud of \textit{banque} is close to the middle part of the point cloud of \textit{bank}.
After removing the translation pairs containing words having the same form or lemma as \textit{bank}, the point cloud of \textit{banque} is moving to the top part of the \textit{bank} point cloud, which is the cluster of the \textit{financial institution} meaning of \textit{bank}.

We take this as meaning that
after removing incomplete supervision data (translation pairs in the dictionary) for multi-sense words, the alignment for multi-sense words is indirectly improved thanks to better supervision data for general embedding spaces mapping.

\subsection{Lexicon Induction Task}
In Table~\ref{tab:eval-result}, we show the accuracy of the lexicon induction task based on different alignments.

For supervised cross-lingual word embedding alignment, we found that removing translation pairs containing words having the same form or lemma as homonym words does not largely affect the lexicon induction task results (around $0.6\%$ difference in the precision at $k=1$).

We observe below the difference between the baseline predictions and the form-based removal predictions (1st LSTM output layer, P@1) in two aspects: 
\begin{itemize}
    \item \textbf{Baseline prediction is correct while the form-based removal prediction is wrong.}
    In this case, we found some of the form-based removal predictions are indeed correct and that the gold standard is incomplete. For instance: 
    \begin{enumerate}
        \item Single-sense word: e.g., \textbf{highlight}, the predicted mapping of the form-based removal is \textit{souligné}, but the gold standard is \textit{souligne}
        \item Multi-sense word: e.g., \textbf{galaxy}, the predicted mapping of the form-based removal is French word \textit{titan}, the gold standard is \textit{galaxie, galaxy}, \textit{galaxy} is a multi-sense word which has the meaning of a group of illustrious people\rempz{the link to \emph{titan} is unclear to me}; \textbf{commands}, the predicted mapping of the form-based removal is \textit{instructions}, the gold standard is \textit{commandements}, \textit{commandes}. \textit{instructions} is another meaning of English word \textit{commands}.
    \end{enumerate}
    \item \textbf{Baseline prediction is wrong while the form-based removal prediction is right.}
    There are 11 words which are aligned correctly by the form-based removal, i.e, \textbf{flute}, gold standard is \textit{fl\^ute}, \textit{fl\^utes}, the baseline method maps it to the French word \textit{trompette}, which is another instrument trumpet; \textbf{madagascar}, the baseline prediction is \textit{mozambique}, the name of one Africa country and also the Mozambique channel between Madagascar and the African mainland.
\end{itemize}

For unsupervised cross-lingual word embedding alignment (Table~\ref{tab:eval-result}), we found that removing exact homonym-related anchor embeddings improves the P@top1 by 10 points and the P@top5 and P@top10 by 5 points (anchors removal(en)).
Removing noisy information about multi-sense words is therefore very beneficial in this case.
Replacing multi-sense word average anchor embeddings with cluster-level average anchors embeddings achieves the best result by using 1st LSTM output layer of ELMo.

\section{Conclusion}
In this paper, we explored the contextual word embeddings (token embeddings) of multi-sense words, argued that the current state-of-the-art method for cross-lingual token embedding learning cannot handle multi-sense words well and proposed our solutions by considering multi-sense word token embeddings as noise. Experiments showed that our methods can improve the token embeddings alignment for multi-sense words in a microscopic perspective without hurting the macroscopic performance on the bilingual lexicon induction task.
As the research on cross-lingual token embedding learning is still in its early stage, we also discussed possible future work such as applying clustering algorithms on token embeddings to obtain sense-level multi-sense word representations.

Possible extensions would be to train a multi-sense word detector based on the number of clusters of token embeddings for each word and to create a new evaluation task for cross-lingual contextual word embeddings (token embeddings) with attention to multi-sense words.

\newpage
\bibliographystyle{dinat}
\bibliography{references}

\newpage
\section*{Appendix}
\begin{table*}[htbp!]
\centering
\resizebox{1.0\textwidth}{!}{
\begin{tabular}{|l|l|p{10cm}|p{10cm}|}
\hline
Word & \begin{tabular}[c]{@{}l@{}}Cluster\\ Positions\end{tabular} & Sentences @ 1st layer & Sentences @ 2nd layer \\ \hline

\multirow{4}{*}{bank} & \multirow{2}{*}{left}  
& Small Craft Company USMC assisted in locating the bodies of the slain snipers and were engaged in a large fire fight on the east \textbf{bank} of the Euphrates River in the city of Haditha.       
& At the northern \textbf{bank} of the Svir River () the Finnish army had prepared a defence in depth area which was fortified with strong-points with concrete pillboxes, barbed wire, obstacles and trenches.
\\ \cline{3-4} && 
The population on the east \textbf{bank} of the Weser had not prepared adequate defenses, so the crusading army attacked there first, massacring most of the population; the few survivors were burnt at the stake.
& These specimens were collected at the Karagachka locality (locality 34 or PIN 2973), to the opposite \textbf{bank} of the Karagatschka River from Karagachka village located in a drainage basin of left bank of the Ural River, Solâ Iletsk district of Orenburg Region, southern European Russia.
\\ \cline{2-4} & \multirow{2}{*}{right} 
& If government bonds that have come due are held by the central \textbf{bank}, the central bank will return any funds paid to it back to the treasury. 
& Issue \textbf{bank} notes;                  
\\ \cline{3-4} && 
Liz is astonished when the police suddenly arrive at the pub to tell her that Jim has been caught robbing a \textbf{bank} and now has a number of hostages.
& Although such measures were not effected, the new administration was successful in tackling other issues: both deficit and the cost of living dropped while the \textbf{bank} reserves trebled, and some palliatives were introduced in lieu of a land reform (the promised tax cuts, plus the freeing of "mainmorte" property).
\\ \hline

\multirow{8}{*}{spring} & \multirow{2}{*}{\begin{tabular}[c]{@{}l@{}}top \\ left\end{tabular}} 
& However, after reaching Ulster the horse stops and urinates, and a \textbf{spring} rises from the spot.
& The \textbf{spring} had been shut off by a rock 74 meters long and 30 meters wide, which obstructed the construction of a running water system. \\ \cline{3-4} &  
 & Over running water â Literally "living", that is, \textbf{spring} water.
 & The holy \textbf{spring} is known to change its colour with various hues of red, pink, orange, green, blue, white, etc. \\ \cline{2-4} 
 & \multirow{2}{*}{\begin{tabular}[c]{@{}l@{}}bottom \\ left\end{tabular}} 
 & A 5'10", 170-pound infielder, Werber was at \textbf{spring} training and toured for several weeks in July with the Yankees in 1927.
 & Joss attended \textbf{spring} training with Cleveland before the start of the 1911 season. \\ \cline{3-4} &  
 & He was invited to \textbf{spring} training and sent to minor league camp on March 14.
 & He pitched in the California Angels minor league system in the early 1990s and participated in "Replacement player" \textbf{spring} training games in 1995 for the Toronto Blue Jays. \\ \cline{2-4} 
 & \multirow{2}{*}{\begin{tabular}[c]{@{}l@{}}bottom\\ middle\end{tabular}} 
 & In \textbf{spring} 912, the Jin attack against Yan got underway, with Zhou commanding the Jin army in a joint operation with the Zhao general Wang Deming (Wang Rong's adoptive son) and the Yiwu Circuit (headquartered in modern Baoding, Hebei) army commanded by Cheng Yan (whose military governor, Wang Chuzhi, was also a Jin ally).
 & In \textbf{spring} 2017, Ponders hit the road supporting Pouya and Fat Nick, opening to sellout crowds across Ontario and Quebec. \\ \cline{3-4} &  
 & In \textbf{spring} 2010 CSX railroad removed the diamonds connecting the southern portion of the Belt Railroad, thus isolating the line from the U.S. rail system.
 & In \textbf{spring} 1944, the Rabstejn sub-camp of Flossenburg was created here, with a capacity of 600 prisoners. \\ \cline{2-4} 
 & \multirow{2}{*}{right} 
 & In the \textbf{spring} of 1935, the All-Union Organization of Cultural Relations with Foreign Countries agreed to send a delegation to the upcoming First International Festival of the Folk Dance in London.
 & Hirsig's role as Crowley's initiatrix reached a pinnacle in the \textbf{spring} of 1921 when she presided over his attainment of the grade of Ipsissimus, the only witness to the event. \\ \cline{3-4} &  
 & In the \textbf{spring} of 2012 in Pakistan was established Pakistani mission.
 & Brown wrote, "In the \textbf{spring} of 1819 a nightingale had built her nest near my house. \\ \hline
 
 \multirow{4}{*}{check} & \multirow{2}{*}{left}  
& Because the defined cases are exhaustive, the compiler can \textbf{check} that all cases are handled in a pattern match:       
& It is standardized for use by mud engineers to \textbf{check} the quality of drilling mud.                                     
\\ \cline{3-4} &
& Most spotters maintained books of different aircraft fleets and would underline or \textbf{check} each aircraft seen.
& The lowest level, where the sounds are the most fundamental, a machine would \textbf{check} for simple and more probabilistic rules of what sound should represent.
\\ \cline{2-4} & \multirow{2}{*}{right} 
& Usually, the trial \textbf{check} will quickly reject the trial match. 
& It is important to realize that glucose-glutamic acid is not intended to be an accuracy \textbf{check} in the test.\\ \cline{3-4} &
& The donor's hematocrit or hemoglobin level is tested to make sure that the loss of blood will not make them anemic, and this \textbf{check} is the most common reason that a donor is ineligible. 
& U.S. Attorney General John Mitchell, citing an extensive background \textbf{check} by the Justice Department, was willing to forgive, stating that it was unfair to criticize Carswell for "political remarks made 22 years ago." 
\\ \hline
 
 \multirow{6}{*}{clear} & \multirow{2}{*}{top} 
 & From here, she had to fight an uphill battle to \textbf{clear} her name and proved her right by finding the authentic painting, while she was also struggling with financial hardship and interference from Min Jung-hak. 
 & On 1 November, Ouagadougou Mayor Simon CompaorÃ© led volunteers on "Operation Mana Mana" (Operation Clean-Clean in Dyula) to \textbf{clear} the streets, which earned him praise on social media. \\ \cline{3-4} 
 &  
 & Jones' shoulder injury came after Botha attempted to \textbf{clear} him from a ruck and the Bulls star was subsequently cited and banned for two weeks for the challenge. 
 & Again a gold medal favourite in the 110 metre hurdles at the London Olympics he pulled his Achilles tendon attempting to \textbf{clear} the first hurdle in the heats. \\ \cline{2-4} 
 & \multirow{2}{*}{\begin{tabular}[c]{@{}l@{}}Bottom \\ left\end{tabular}} 
 & She made it \textbf{clear} that she did not intend for Nassar to ever be free again. 
 & Hugenberg for his part regarded "Katastrophenpolitik" as a good idea that was unfortunately abandoned, and made it \textbf{clear} that he wanted a return to "Katastrophenpolitik". \\ \cline{3-4} 
 &  
 & Many Southerners felt that the Compromise of 1850 had been shaped more towards Northern interests; the Georgia Platform made it \textbf{clear} that the future of the nation depended on the North strictly adhering to the Compromise. 
 & The political heat was turned up on the issue since Bush mentioned changing Social Security during the 2004 elections, and since he made it \textbf{clear} in his nationally televised January 2005 speech that he intended to work to partially privatize the system during his second term. \\ \cline{2-4} 
 & \multirow{2}{*}{\begin{tabular}[c]{@{}l@{}}Bottom\\ right\end{tabular}} 
 & However, in "Reference re Secession of Quebec", the Supreme Court of Canada has essentially said that a democratic vote in itself would have no legal effect, since the secession of a province in Canada would only be constitutionally valid after a negotiation between the federal government and the provincial government; whose people would have clearly expressed, by a \textbf{clear} majority, that it no longer wished to be part of Canada. 
 & He was the \textbf{clear} winner with ten seconds over the runner-up, fellow Kenyan Albert Kiptoo Yator. \\ \cline{3-4} 
 &  & The game sees Kasparov rejecting \textbf{clear} drawing opportunities and eventually losing. 
 & He wrote to Irene Tasker in South Africa, in a \textbf{clear} hand, telling her how much better he was. \\ \hline
\end{tabular}
}
\caption[Corresponding sentences selected from the token embedding clusters of the English words \textit{bank}, \textit{spring}, \textit{check} and \textit{clear}]
{Corresponding sentences selected from the token embedding clusters of the English words \textit{bank}, \textit{spring}, \textit{check} and \textit{clear}.}
\label{table:multi-word-sentences}
\end{table*}

\end{document}